\documentclass[sigconf,screen]{acmart}
\usepackage{url}
\usepackage{amsthm}
\usepackage{afterpage}
\usepackage{subcaption}
\usepackage[normalem]{ulem}
\useunder{\uline}{\ul}{}
\usepackage{graphicx}
\usepackage{textcomp}
\usepackage{xcolor}
\usepackage{balance} 
\usepackage{bbm}
\usepackage{amsmath,amsfonts}
\usepackage{booktabs}
\usepackage{tcolorbox}

\usepackage{epstopdf}
\usepackage{array}
\usepackage{booktabs}
\usepackage{multicol}
\usepackage{color}
\usepackage{xcolor}
\usepackage{colortbl}
\usepackage{xspace}
\usepackage{multirow}
\usepackage{pifont}
\usepackage{enumitem}
\usepackage{bbding}
\usepackage{hyperref}
\usepackage{makecell}
\usepackage{microtype}
\usepackage{fontawesome}
\usepackage{caption,setspace}
\usepackage[utf8]{inputenc}
\usepackage{utfsym}
\usepackage{diagbox}
\usepackage{makecell}
\newcommand{\thickhline}{\noalign{\hrule height 0.3pt}} 
\DeclareMathOperator*{\argmax}{arg\,max}
\DeclareMathOperator*{\argmin}{arg\,min}

\AtBeginDocument{%
 }

\copyrightyear{2025}
\acmYear{2025}
\setcopyright{acmlicensed}
\acmConference[KDD '26]{Proceedings of the 32st ACM SIGKDD Conference on Knowledge Discovery and Data Mining V.2}{August 9--12, 2026}{Jeju, Korea}
\acmBooktitle{Proceedings of the 31st ACM SIGKDD Conference on Knowledge Discovery and Data Mining V.2 (KDD '26), August 9--12, 2026, Jeju, Korea}
\acmDOI{10.1145/3770855.3817900}
\acmISBN{979-8-4007-1454-2/2025/08}



\usepackage[linesnumbered,ruled,vlined]{algorithm2e}

\newcommand{\ssymbol}[1]{^{\@fnsymbol{#1}}}

\SetKwInput{KwInput}{Input}                
\SetKwInput{KwOutput}{Output}              
\SetKwRepeat{Do}{do}{while}

\definecolor{darkred}{rgb}{0.55, 0.0, 0.0}
\definecolor{lightred}{rgb}{0.94, 0.5, 0.5}
\definecolor{rosybrown}{rgb}{0.74, 0.56, 0.56}
\definecolor{darkgreen}{rgb}{0.0, 0.39, 0.0}
\definecolor{skyblue}{rgb}{0.56, 0.93, 0.56}
\definecolor{royalblue}{rgb}{0.0, 0.0, 0.55}
\definecolor{lightblue}{rgb}{0.68, 0.85, 0.9}
\definecolor{skyblue}{rgb}{0.53, 0.81, 0.92}

\definecolor{gray}{rgb}{0.5, 0.5, 0.5}
\definecolor{forestgreen}{rgb}{0.13, 0.55, 0.13}
\definecolor{slateblue}{rgb}{0.42, 0.35, 0.80}
\definecolor{darkgray}{rgb}{0.33, 0.33, 0.33}
\definecolor{royalblue}{rgb}{0.25, 0.41, 0.88}

\SetCommentSty{mycommfont}

\usepackage{ulem}

\begin{document}

\title{Both Topology and Text Matter: Revisiting LLM-guided Out-of-Distribution Detection on Text-attributed Graphs}

\author{Yinlin Zhu}
\affiliation{%
\institution{Sun Yat-sen University}
\city{Guangzhou}
\country{China}}
\email{zhuylin27@mail2.sysu.edu.cn}

\author{Di Wu}
\authornote{Corresponding author}
\affiliation{
  \institution{Sun Yat-sen University}
  \city{Guangzhou}
  \country{China}
}
\email{wudi27@mail.sysu.edu.cn}

\author{Xu Wang}
\affiliation{%
\institution{Shandong University}
\city{Weihai}
\country{China}}
\email{xuwang2023@mail.sdu.edu.cn}

\author{Guocong Quan}
\affiliation{%
\institution{Sun Yat-sen University}
\city{Guangzhou}
\country{China}}
\email{quangc@mail.sysu.edu.cn}

\author{Miao Hu}
\affiliation{%
\institution{Sun Yat-sen University}
\city{Guangzhou}
\country{China}}
\email{humiao5@mail.sysu.edu.cn}

\renewcommand{\shortauthors}{Yinlin Zhu, Di Wu, Xu Wang, Guocong Quan, Miao Hu}

\begin{abstract}
Text-attributed graphs (TAGs) associate nodes with textual attributes and graph structure, enabling GNNs to jointly model semantic and structural information. Although effective on in-distribution (ID) data, GNNs often fail on out-of-distribution (OOD) nodes with unseen textual or structural patterns, producing overconfident predictions without reliable OOD detection. Existing topology-driven methods mitigate node-level bias through neighboring structures, but typically encode texts as shallow features, underutilizing semantic information. Recent LLM-based approaches instead synthesize pseudo OOD priors from textual knowledge, yet suffer from two key limitations: (1) a trade-off between reliability and informativeness, where generated OOD exposures either deviate from true OOD semantics or introduce substantial ID noise; and (2) dependence on specialized architectures, limiting compatibility with topology-level advances validated in prior work.
To address these issues, we propose \textsc{\textbf{LG-Plug}}, an \underline{\textbf{L}}LM-\underline{\textbf{G}}uided \underline{\textbf{\textsc{Plug}}}-and-play framework for TAG OOD detection. \textsc{LG-Plug} aligns topology and text representations to obtain fine-grained node embeddings, then constructs consensus-driven OOD exposure through clustered iterative LLM prompting. To reduce LLM query cost, it further adopts lightweight in-cluster codebooks and heuristic sampling. The generated OOD exposure acts as a regularizer that separates ID and OOD nodes, enabling seamless integration with existing detectors.
Experiments on six TAG benchmarks demonstrate that \textsc{LG-Plug} consistently improves topology-driven OOD detectors ($\geqslant$7\% FPR95 reduction) and surpasses prior LLM-based methods ($\geqslant$5\% FPR95 reduction).
\end{abstract}

 \maketitle

\section{Introduction}
\label{sec: introduction}

Text-attributed graphs (TAGs) represent relational data where each node is associated with a textual description and each edge denotes an interaction between entities~\cite{sharma2024tag_realworld}. As a powerful graph learning paradigm, graph neural networks (GNNs) leverage message passing to jointly model the knowledge encoded in node texts and graph topology, achieving strong performance on a variety of TAG benchmarks~\cite{zhao2024tag_node_cls_fewshot, yu2025llm4tag_node_gen_fewshot, zhang2025llm4tag_label_free_nodecls}. Despite these successes, GNNs remain highly sensitive to distribution shifts in realistic deployment scenarios. In particular, although trained on labeled in-distribution (ID) nodes drawn from a pre-defined label space, GNNs are often required to handle textual descriptions or structural patterns whose semantics lie far beyond the training distribution, such as newly emerging attack patterns in threat-intelligence graphs~\cite{sarhan2021tag_opencykg} and previously unreported clinical conditions in biomedical citation graphs~\cite{sen2008tag_dataset_cora_citeseer}. Without robust out-of-distribution (OOD) detection mechanisms, GNNs tend to produce overconfident yet incorrect predictions, posing substantial risks in safety-critical applications~\cite{gui2022graph_ood_benchmark}.

To mitigate these risks, OOD detection on graphs has been widely studied in recent years. A substantial body of prior work focuses on topology-level signals, aiming to characterize the OODness of a target node through the OODness of its topological neighbors. Representative approaches include topology-aware energy propagation~\cite{wu2023graph_ood_gnnsafe, yang2025graph_ood_nodesafe}, topology augmentation~\cite{ma2024graph_ood_grasp}, and architectural modeling of OOD graph structures~\cite{wang2025graph_ood_gold}. Compared with non-graph baselines such as MSP~\cite{Hendrycks2017MSP} and ODIN~\cite{liang2018odin}, these methods typically achieve notable performance improvements, underscoring the effectiveness of their topology-level insights. However, most of these approaches were developed when language modeling capabilities were relatively limited~\cite{cai2025graph_ood_survey} and thus treat node texts as shallow features, such as bag-of-words representations~\cite{Yang16cora, sen2008tag_dataset_cora_citeseer}. As a result, these methods fail to fully exploit the semantic knowledge of node textual descriptions in TAGs, leading to limited discriminability between ID and OOD nodes.

Recently, the rapid progress of large language models (LLMs) has reshaped this research landscape. Trained on massive and diverse corpora, LLMs such as GPT-4~\cite{achiam2023gpt4} demonstrate strong contextual understanding and semantic generalization, enabling more effective integration of textual semantics for TAG-oriented OOD detection. Building upon this insights, most existing approaches employ LLMs to synthesize pseudo OOD labels or corresponding samples, essentially performing pseudo OOD exposure through LLM-generated semantic priors. For example, LLMGuard~\cite{lv2025tag_ood_llm_llmguard} generates multiple pseudo OOD labels together with associated synthetic samples conditioned on ID class labels, and further adopts instruction tuning to integrate their potential topological associations. In addition, GLIP-OOD~\cite{xu2025zero_shot_tag_ood_glip} and GOE-LLM~\cite{xu2025goe_llm} further exploit the transductive setting by randomly sampling unlabeled nodes from the graph and prompting LLMs to assign pseudo OOD labels, which are then utilized for subsequent downstream training.

\begin{figure*}[h]
    \includegraphics[width=0.998\textwidth]{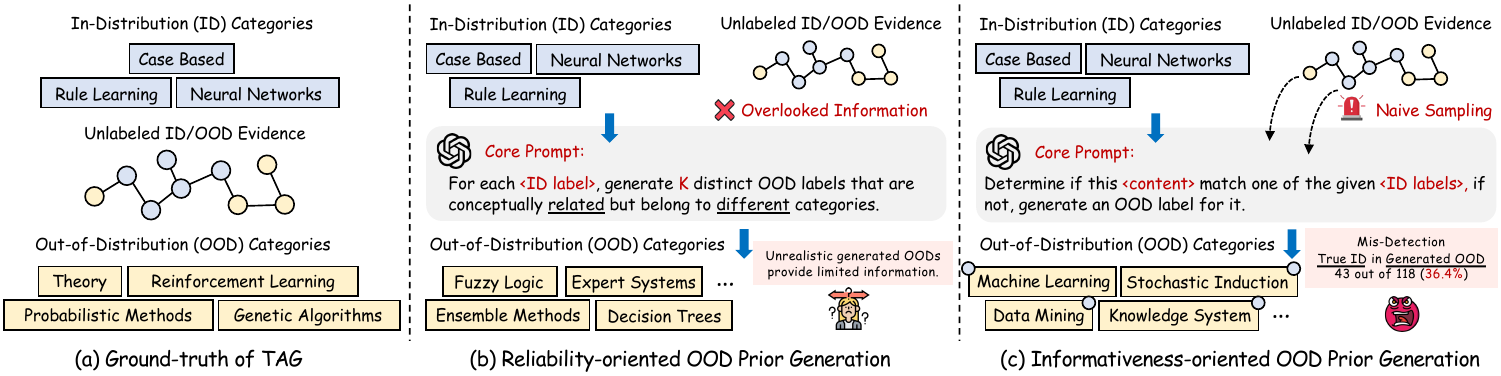}
    \captionsetup{font={small,stretch=1}}
    \caption{Illustration of the reliability-informativeness imbalance in LLM-based OOD prior generation:
(a) Realistic ID and OOD categories on Cora~\cite{Yang16cora}.
(b) Reliability-oriented methods avoid ID overlap but yield semantically unrealistic priors, limiting informativeness.
(c) Informativeness-oriented methods produce realistic OOD priors by labeling unlabeled nodes but risk mislabeling IDs to harm detection.}
    \label{fig: illustration_existing_llm_ood}
\end{figure*}

Despite promising results, existing LLM-based OOD detectors suffer from two key limitations. (1) \textbf{Reliability-informativeness imbalance:} From the perspective of \textcolor{darkred}{\textbf{Effectiveness}}, existing LLM-based OOD prior generation methods face a trade-off between reliability and informativeness. Reliability-oriented methods (Fig.~\ref{fig: illustration_existing_llm_ood} (b)), such as LLMGuard~\cite{lv2025tag_ood_llm_llmguard}, generate OOD priors by prompting LLMs with ID class labels, ensuring samples do not overlap with ID data but often producing semantically unrealistic OOD examples. In contrast, informativeness-oriented methods (Fig.~\ref{fig: illustration_existing_llm_ood} (c)), such as GLIP-OOD~\cite{xu2025zero_shot_tag_ood_glip} and GOE-LLM~\cite{xu2025goe_llm}, rely on randomly sampling unlabeled nodes from the original graph to generate OOD priors closer to realistic semantics, but risk mislabeling true ID nodes, introducing noise in downstream training. (2) \textbf{Limited compatibility with topology-driven OOD methods:} From the perspective of \textcolor{royalblue}{\textbf{Compatibility}}, these approaches rely on bespoke end-to-end architectures that tightly couple semantic modeling with the detector, making them difficult to integrate with the broader ecosystem of topology-driven OOD methods that are validated effective.

Building upon these insights, we pose the following questions:
\textbf{Q1}: How can LLMs be leveraged to generate pseudo OOD priors that are both reliable and informative?
\textbf{Q2}: How can LLM-derived OOD priors be effectively utilized while remaining compatible with existing topology-driven graph OOD detectors?

To this end, we propose \textsc{\textbf{LG-Plug}}, which is an \underline{\textbf{L}}LM-\underline{\textbf{G}}uided \underline{\textsc{\textbf{Plug}}}-and-play strategy for TAG OOD detection tasks. To effectively address \textbf{Q1}, the proposed \textsc{LG-Plug} first performs a topology-text representation alignment, thereby producing fine-grained node embeddings that induce an initial separation between ID and OOD nodes. Based on this alignment, we further design a consensus-driven OOD exposure scheme. Instead of generating OOD priors directly from ID labels or random sampled individual node descriptions, we cluster and filter nodes in the learned embedding space, then progressively capture the consistent semantics within each cluster through iterative LLM prompting for OOD exposure. Moreover, to ensure efficiency, \textsc{LG-Plug} significantly minimizes the querying of LLM by maintaining a lightweight in-cluster category codebook together with a simple heuristic sampling strategy. To address \textbf{Q2}, The resulting OOD exposure is used as a regularization term to further separate ID and OOD nodes, enabling seamless integration with existing topology-driven graph OOD detectors.

\textbf{Our Contributions:}
(1) \textbf{Valuable Insights.} We identify two major limitations of existing LLM-based OOD detection methods for TAGs including reliability-informativeness imbalance and limited compatibility with topology-driven OOD methods.
(2) \textbf{Novel Method.} We propose \textsc{LG-Plug}, which uses topology-text representation alignment and consensus-driven LLM OOD exposure to generate high-quality OOD priors. These priors can serve as a regularization term, seamlessly integrating with most topology-driven graph OOD methods to enhance their performance.
(3) \textbf{State-of-the-art Performance.} Extensive experiments on six benchmark TAG datasets show that \textsc{LG-Plug} consistently improves a variety of topology-driven OOD methods ($\geqslant$7\% FPR95 reduction) and achieves state-of-the-art results compared with LLM-based graph OOD baselines ($\geqslant$5\% FPR95 reduction).

\vspace{-0.05cm}
\section{Related Works}
\label{sec: related works}

\noindent \textbf{Graph Neural Networks.}
Earlier research on deep graph learning extends convolution to handle graphs~\cite{bruna2014spectral} but comes with notable parameter counts. To this end, GCN~\cite{kipf2016gcn} simplifies graph convolution by utilizing a 1-order Chebyshev filter to capture local neighborhood information.  GAT~\cite{velivckovic2017gat} adopts graph attention, allowing weighted aggregation. GraphSAGE~\cite{hamilton2017graphsage} introduces a variety of learnable aggregation functions for performing message aggregation. Moreover, GIN~\cite{xu2018gin} aims to preserve structural information maximally and proves its discriminative power matches the Weisfeiler-Lehman graph isomorphism test. Further details on GNN research can be found in surveys~\cite{wu2020gnn_survey1, zhou2022gnn_survey2}.

\vspace{+0.1cm}
\noindent \textbf{LLM for Text-Attributed Graphs.} Text-attributed graphs (TAGs), characterized by node-level textual descriptions and edge-level entity interactions, serve as a universal framework for representing relational data across diverse domains~\cite{feng2024tag_taglas}. Driven by the advancements in large language models (LLMs) such as GPT-4~\cite{achiam2023gpt4}, recent research has extensively explored LLM-based methodologies to enhance graph-related tasks~\cite{yu2025llm4tag_node_gen_fewshot, zhang2025llm4tag_label_free_nodecls}. These efforts can be categorized into four primary paradigms: (1) GNN as Prefix, where graph data is encoded as token sequences to enable efficient processing by LLMs, as seen in GraphGPT~\cite{tang2024gnns_as_prefix_graphgpt}, Unigraph~\cite{he2024gnns_as_prefix_unigraph}, and GraphLLM~\cite{chai2025gnns_as_prefix_graphllm}; (2) LLMs as Prefix, where LLMs first extract node embeddings or generate labels from textual metadata to facilitate the subsequent training of graph neural networks, exemplified by SimTeG~\cite{duan2023llms_as_prefix_simteg}, OFA~\cite{liu2023llms_as_prefix_ofa}, and LLM-GNN~\cite{chen2023llms_as_prefix_llms_gnn}; (3) LLM-Graph Integration, which involves deep alignment, fusion training, or the development of graph-interactive agents, including G2P2~\cite{wen2023llms_graphs_intergration_g2p2}, GLEM~\cite{zhao2022llms_graphs_intergration_glem}, and ENGINE~\cite{zhu2024llms_graphs_intergration_engine}; and (4) LLMs-Only, which focuses on sophisticated prompt engineering to enable LLMs to reason over graph structures directly, as demonstrated in InstructGLM~\cite{ye2024llms_only_instructGLM}, GPT4Graph~\cite{guo2023llms_only_gpt4graph}, and NLGraph~\cite{wang2023llms_only_nlgraph}. Further details can be found in surveys~\cite{su2025llm4tag_survey, ren2024llm_for_tag_survey_2}.

\vspace{+0.1cm}
\noindent \textbf{Out-of-Distribution Detection on Graphs.}
Out-of-distribution (OOD) detection aims to identify test samples whose attributes or labels deviate from the training distribution, thereby improving model reliability under distribution shifts~\cite{gui2022graph_ood_benchmark}. Early OOD detection studies primarily focus on non-graph settings, including confidence-based and energy-based methods such as MSP~\cite{Hendrycks2017MSP}, ODIN~\cite{liang2018odin}, and energy-based scoring~\cite{liu2020energy}. However, directly extending them to graphs often yields suboptimal performance, as they fail to exploit topology that is critical for characterizing node OODness.

To address this limitation, a line of topology-driven graph OOD detection methods has been proposed. GKDE~\cite{zhao2020uncertainty} models node-level uncertainty by estimating Dirichlet distributions over GNN outputs, enabling OOD detection via multi-source uncertainty estimation. GNNSAFE~\cite{wu2023graph_ood_gnnsafe} further propagates energy scores along graph topology to enhance robustness against locally ambiguous predictions, while NodeSAFE~\cite{yang2025graph_ood_nodesafe} introduces energy regularization to suppress extreme confidence values. GRASP~\cite{ma2024graph_ood_grasp} improves OOD detection by performing topology-aware edge augmentation, facilitating more effective OOD score propagation among neighbors. 

Despite their effectiveness, most existing topology-driven methods are developed under limited language modeling assumptions (e.g., bag-of-words) and treat node textual descriptions as shallow features. In text-attributed graphs (TAGs), however, node OODness is jointly determined by textual content and graph structure. Motivated by recent advances in large language models (LLMs), emerging approaches leverage LLMs to synthesize pseudo OOD labels or samples for TAG-oriented OOD detection. LLMGuard~\cite{lv2025tag_ood_llm_llmguard} generates pseudo OOD labels and corresponding synthetic nodes by prompting LLMs with ID class labels under domain constraints, ensuring reliable separation from ID classes. In contrast, GLIP-OOD~\cite{xu2025zero_shot_tag_ood_glip} and GOE-LLM~\cite{xu2025goe_llm} exploit the transductive setting by randomly sampling unlabeled nodes and prompting LLMs to infer potential OOD semantics, yielding more informative but potentially noisy OOD priors. While promising, these LLM-based approaches still face challenges in effectiveness and compatibility with existing topology-driven graph OOD detectors, as introduced in Sec.~\ref{sec: introduction}.

\vspace{+0.2cm}
\section{Problem Formulization}
\label{sec: problem formulization}

We focus on semi-supervised, node-level OOD detection in TAGs. For a given TAG $G=(\mathcal{V}, \mathcal{E})$, $\mathcal{V}$ denotes the set of nodes and $\mathcal{E}$ is the set of edges. Each node $v_i \in \mathcal{V}$ is associated with a textual description $t_i \in \mathcal{T}$. The set of nodes $\mathcal{V}$ is partitioned into a set of labeled nodes $\mathcal{V}_l$ and an unlabeled set $\mathcal{V}_u$, such that $\mathcal{V} = \mathcal{V}_l \cup \mathcal{V}_u$. Importantly, the unlabeled nodes $\mathcal{V}_u$ further comprise ID nodes $\mathcal{V}_{u}^{id}$ and OOD nodes $\mathcal{V}_{u}^{ood}$, where $\mathcal{V}_u = \mathcal{V}_{u}^{id} \cup \mathcal{V}_{u}^{ood}$. The label space for ID nodes is $\mathcal{Y}_{id} = \{y_1, y_2, \ldots, y_C\}$, with $C$ being the number of ID classes. In contrast, the OOD label space $\mathcal{Y}_{ood}$ is unknown. Let $\mathcal{D}_{id}$ denote the distribution of ID nodes $v \in \mathcal{V}_{u}^{id}$ with labels $y \in \mathcal{Y}_{id}$, and $\mathcal{D}_{ood}$ represent the OOD distribution, consisting of nodes $v \in \mathcal{V}_{u}^{ood}$ with unmodeled semantics and unknown labels $y \in \mathcal{Y}_{ood}$. Graph OOD detection aims to determine whether a given unlabeled node $v_i \in \mathcal{V}_u$ belongs to the $\mathcal{D}_{id}$ or $\mathcal{D}_{ood}$ class.

\section{Methodology}
\label{sec: methodology}

\begin{figure*}[h]
    \includegraphics[width=0.998\textwidth]{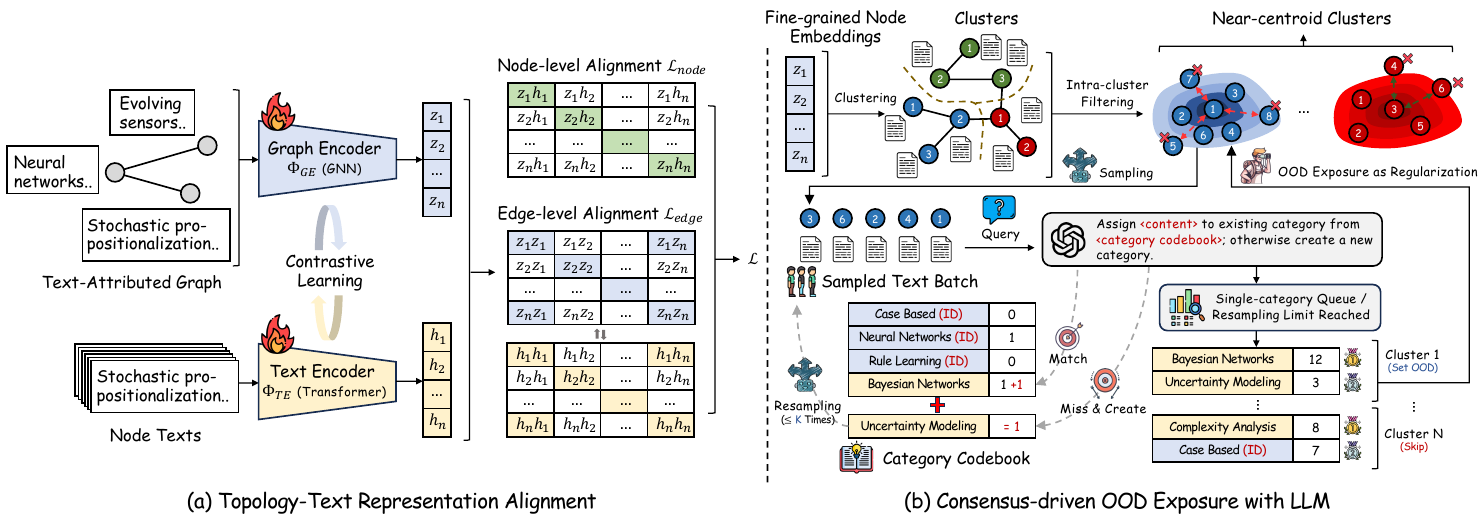}
    \captionsetup{font={small,stretch=1}}
    \caption{Overview of the proposed \textsc{LG-Plug} plug-and-play framework. (a) \textit{Topology-Text Representation Alignment} jointly trains a graph encoder and a text encoder to learn fine-grained and discriminative node embeddings for both ID and OOD nodes. (b) \textit{Consensus-driven OOD Exposure with LLM} identifies reliable and informative OOD exposures from unlabeled nodes by prompting a LLM while remaining scalable. The extracted OOD exposures act as a regularization signal and can be seamlessly integrated into existing topology-driven graph OOD detectors.}
    \label{fig: framework_lgplug}
\end{figure*}

In this section, we introduce \textsc{LG-Plug}, an LLM-guided plug-and-play strategy for OOD detection on TAGs. \textsc{LG-Plug} is designed to augment existing topology-driven OOD detectors with LLM-derived semantic priors, while preserving their architectural compatibility. An overview of the framework is shown in Fig.~\ref{fig: framework_lgplug}.
The design of \textsc{LG-Plug} directly corresponds to the three challenges and research questions (\textbf{Q1}-\textbf{Q2}) discussed in Sec.~\ref{sec: introduction}.
To address \textbf{Q1} (\textcolor{darkred}{\textbf{Effectiveness}}), \textsc{LG-Plug} aligns topology-aware representations with textual semantics to obtain fine-grained node embeddings, and then applies a consensus-driven LLM annotation scheme that clusters nodes and progressively filters ambiguous samples, yielding reliable and informative OOD exposure. Moreover, to ensure efficiency, \textsc{LG-Plug} avoids reduces time cost of LLM querying through a lightweight in-cluster category codebook and heuristic sampling strategy.
To address \textbf{Q2} (\textcolor{royalblue}{\textbf{Compatibility}}), the generated OOD exposure is incorporated as a lightweight OOD score regularization term, enabling seamless integration with a wide range of existing topology-driven graph OOD detectors.

\subsection{Topology-text Representation Alignment}
\label{sec: method_m1}

To address \textbf{Q1} (\textcolor{darkred}{\textbf{Effectiveness}}), we build on the insights from previous studies~\cite{NEURIPS2023tag_benchmark}, where in TAGs, nodes from different categories exhibit discriminative patterns jointly in textual semantics and local topological contexts. To this end, we first leverage a topology-text representation alignment mechanism to capture fine-grained textual semantics while preserving topology-induced inductive biases, learning distinguishable representations for ID and OOD nodes.

Formally, for a given TAG $G=(\mathcal{V},\mathcal{E})$ with node texts $\mathcal{T}$, we employ a 2-layer graph convolutional network (GCN)~\cite{kipf2016gcn} as the graph encoder $\Phi_{\text{GE}}$ to map each node $v_i$ to a $d$-dimensional embedding:
\begin{equation}
\label{eq: graph embedding}
\mathbf{z}_i = \Phi_{\text{GE}}(v_i \mid G, \mathbf{X}, \theta_{\text{GE}}),
\end{equation}
\noindent where $\mathbf{X}\in\mathbb{R}^{|\mathcal{V}|\times d}$ denotes the initial node feature matrix derived from the corresponding textual descriptions of all nodes in the graph, with each row representing a $d$-dimensional feature vector, and $\theta_{\text{GE}}$ signifies the parameter set of the graph encoder.

In parallel, we adopt a Transformer~\cite{vaswani2017attention_is_all_you_need}-based text encoder $\Phi_{\text{TE}}$ to encode the raw text $\mathcal{T}_i$ of node $v_i$ into a $d$-dimensional representation, which is computed as follows:
\begin{equation}
\label{eq: text embedding}
\mathbf{h}_i = \Phi_{\text{TE}}(\mathcal{T}_i \mid \theta_{\text{TE}}),
\end{equation}
\noindent where $\theta_{\text{GE}}$ signifies the parameter set of the text encoder.

\vspace{+0.1cm}
\noindent \textbf{Node-level Alignment.}
Let $\mathbf{Z}, \mathbf{H}\in\mathbb{R}^{n\times d}$ denote the obtained graph and text embeddings for a batch of $n$ nodes, we first perform node-wise alignment by constructing a similarity matrix:
\begin{equation}
\mathbf{\Lambda}1 = (\tilde{\mathbf{Z}}\tilde{\mathbf{H}}^\top)\cdot \exp(\tau),
\end{equation}
where $\tilde{\mathbf{Z}}$ and $\tilde{\mathbf{H}}$ are row-wise $L_2$-normalized embeddings and $\tau$ is a temperature parameter. Then, the alignment objective is defined via a symmetric contrastive loss, formulated as follows:
\begin{equation}
\label{eq: node loss}
\mathcal{L}_{\text{node}} = \frac{1}{2}\big(\text{CE}(\mathbf{\Lambda}_1,\mathbf{y}) + \text{CE}(\mathbf{\Lambda}_1^\top,\mathbf{y})\big),
\end{equation}
where $\mathbf{y}=(1,2,\dots,n)^\top$ denotes the identity labels and $\text{CE}(\cdot)$ is the cross-entropy loss.

\vspace{+0.1cm}
\noindent \textbf{Edge-level Alignment.} Moreover, to further exploit higher-order interactions on the graph, we align the relational proximity between the textual modality and the topological modality. We define $\mathbf{\Lambda}_2 = (\tilde{\mathbf{Z}}\tilde{\mathbf{Z}}^{\top})$ and $\mathbf{\Lambda}_3 = (\tilde{\mathbf{H}}\tilde{\mathbf{H}}^{\top})$ as the intra-modal similarity matrices. For each edge $(v_i, v_j) \in \mathcal{E}$, we minimize the discrepancy between their topological and textual similarities:
\begin{equation}
    \label{eq: edge loss}
    \mathcal{L}_{\text{edge}} = \frac{1}{|\mathcal{E}|} \sum_{(v_i, v_j) \in \mathcal{E}} \| (\mathbf{\Lambda}_2)_{i,j} - (\mathbf{\Lambda}_3)_{i,j} \|^2.
\end{equation}

\noindent \textbf{Overall Alignment Objective.} To derive the optimal graph encoder $\theta_\textbf{GE}$ and text encoder $\theta_\textbf{TE}$, we jointly optimize both of them by minimizing a unified contrastive loss that integrates node-level and edge-level alignments:
\begin{equation}
    \label{eq: ttra obj}
    \theta_\text{GE}^*, \theta_\text{TE}^* = \mathop{\arg\min}_{\theta_\text{GE}, \theta_\text{TE}} \mathcal{L}_{\text{node}} + \lambda \, \mathcal{L}_{\text{edge}},
\end{equation}
where $\lambda$ is a trade-off hyperparameter that balances the importance of node-level alignment and edge-level alignment.

\subsection{Consensus-driven OOD Exposure with LLM}
\label{sec: method_m2}

To further address \textbf{Q1} (\textcolor{darkred}{\textbf{Effectiveness}}), after obtaining the topology-text aligned node representation, we identify reliable and informative OOD from unlabeled nodes. A key challenge is that OODness cannot be robustly determined at the individual node level, as semantic deviation from ID classes lacks a clear decision boundary. In contrast, when multiple nodes share consistent non-ID semantics, their collective patterns provide stronger evidence for defining OOD concepts. Motivated by this intuition, we propose a consensus-driven LLM annotation scheme that identifies OOD exposure at the cluster level rather than the instance level.

\vspace{+0.1cm}
\noindent \textbf{Clustering-based Candidate Grouping.}
Let $\mathbf{Z}\in\mathbb{R}^{|\mathcal{V}|\times d}$ denote the aligned node embeddings.
We first perform clustering in the embedding space to partition unlabeled nodes into $M$ clusters, each expected to group nodes with similar semantic and topological characteristics. Formally, we solve the following optimization problem:
\begin{equation}
\label{eq: clustering}
\{\mathcal{C}_m\}_{m=1}^M
= \argmin_{{\mathcal{C}_m}}
\sum_{m=1}^M \sum_{v_i \in \mathcal{C}_m}
|| \mathbf{z}_i - \boldsymbol{\mu}_m ||_2^2,
\end{equation}
where $\mathcal{C}_m$ denotes the $m$-th cluster and $\boldsymbol{\mu}_m = \frac{1}{|\mathcal{C}_m|}\sum_{v_i \in \mathcal{C}_m}\mathbf{z}_i$ is the centroid of cluster $\mathcal{C}_m$. In this paper, we approximately solve Eq.~(\ref{eq: clustering}) using the standard K-means algorithm~\cite{lloyd1982kmeans}. Moreover, for each cluster $\mathcal{C}_m$, we further select a small set of representative nodes that are closest to the cluster centroid:
\begin{equation}
\label{eq: near-centroid cluster}
\hat{\mathcal{C}}_m
=
\argmin_{\mathcal{S} \subseteq \mathcal{C}_m, |\mathcal{S}| = \lceil \rho |\mathcal{C}_m|\rceil}
\sum_{v_i \in \mathcal{S}} \left\| \mathbf{z}_i - \boldsymbol{\mu}_m \right\|_2 ,
\end{equation}
where $\hat{\mathcal{C}}_m$ contains the nodes closest to the cluster centroid $\boldsymbol{\mu}_m$, and $\rho \in (0,1)$ is the selection ratio. This representative selection increases semantic coherence within each cluster, facilitating reliable consensus-driven OOD identification.

\vspace{+0.1cm}
\noindent \textbf{Sequential LLM Querying with Consensus Filtering.} For each near-centroid cluster $\mathcal{\hat{C}}_m$, we first initialize a lightweight category codebook, which is formulated as follows:
\begin{equation}
\label{eq: codebook init}
\mathcal{B}_m = \{(c_i, n_i)\}_{c_i\in\mathcal{Y}_{id}},
\end{equation}
where $n_i=0$ is its initial count. The codebook $\mathcal{B}_m$ is dynamically updated during the following LLM querying to track the semantic consensus among representative nodes.

Moreover, to ensure the efficiency of the entire plugin, given a near-centroid cluster $\hat{\mathcal{C}}_m$, we avoid exhaustively querying all nodes in the cluster. 
Instead, we iteratively sample a small batch of representative nodes
$\mathcal{R}_m \subset \hat{\mathcal{C}}_m$ ($|\mathcal{R}_m|=b$ holds, and $b$ denotes the batch size) and query the LLM in a sequential manner.
For each node $v_i \in \mathcal{R}_m^b$, we construct a structured prompt (see \cite{LGPLUG2026}) that consists of its textual description $\mathcal{T}_i$ together with the current category codebook $\mathcal{B}_m$.
The prompt instructs the LLM to either assign $v_i$ to an existing category in $\mathcal{B}_m$ or create a new OOD category:
\begin{equation}
\label{eq: llm query}
\hat{c}_i = \text{LLM}\big(\mathcal{T}_i, \mathcal{B}_m\big),
\end{equation}
\noindent where $\hat{c}_i$ is the category assigned to node $v_i$. 
After assigning $\hat{c}_i$ to node $v_i$, the codebook is updated as:
\begin{equation}
\label{eq: update codebook}
\mathcal{B}_m \gets
\begin{cases}
\{(c_j, n_j + 1)\} \cup \mathcal{B}_m \setminus \{(c_j, n_j)\}, & \text{if } \hat{c}_i = c_j, (c_j, n_j) \in \mathcal{B}_m; \\
\mathcal{B}_m \cup \{(\hat{c}_i, 1)\}, & \text{otherwise}.
\end{cases}
\end{equation}
The LLM querying process is repeated until either all nodes in a batch are assigned to the same category, indicating strong semantic consensus, or a maximum number of trials $T$ is reached.  

After termination, we select the top-$K$ items from the codebook $\mathcal{B}_m = \{(c_i, n_i)\}$ according to their match times:
\begin{equation}
\label{eq: top-K items}
\mathcal{\hat{B}}_m = \argmax_{\substack{\mathcal{S} \subset \mathcal{B}_m, |\mathcal{S}| = K}} \sum_{(c_i,n_i) \in \mathcal{S}} n_i.
\end{equation}
Let $\mathcal{Y}_{\text{top}}$ denote the set of top-$K$ categories selected from the filtered codebook $\hat{\mathcal{B}}_m$.  
The annotations for cluster $\hat{\mathcal{C}}_m$ are regarded as reliable if these categories exclude all ID classes, i.e.,
$\mathcal{Y}_{\text{top}} \cap \mathcal{Y}_{id} = \emptyset$.
In this case, nodes assigned to categories in $\mathcal{Y}_{\text{top}}$ are added to the OOD exposure set (initialized as $\emptyset$):
\begin{equation}
\label{eq: ood exposure}
\mathcal{V}_{\text{exp}} \gets \mathcal{V}_{\text{exp}} \;\cup\;
\{ v_i \in \hat{\mathcal{C}}_m \mid \hat{c}_i \in \mathcal{Y}_{\text{top}} \}.
\end{equation}

By enforcing semantic consensus at the cluster level rather than relying on individual node annotations, this procedure effectively suppresses noisy OOD labels arising from ambiguous nodes.

\subsection{Score Regularization via OOD Exposure}
\label{sec: method_m3}

To address \textbf{Q2} (\textcolor{royalblue}{\textbf{Compatibility}}), we incorporate LLM-derived OOD exposure as a lightweight score-level regularization that integrates seamlessly into topology-driven graph OOD detectors without architectural modifications. Specifically, we consider a generic detector that assigns each node $v_i$ an OOD score $s_i$, formulated as:

\begin{equation}
s_i = f_\theta(v_i \mid G),
\end{equation}
where $f_\theta(\cdot)$ denotes a detector-specific scoring function. 
Let $\mathcal{V}_{\text{exp}}$ denote the OOD exposure set obtained from Sec.~\ref{sec: method_m2}, and let $\mathcal{V}_{id}$ denote the labeled ID nodes. Inspired by \textsc{GNNSafe++}~\cite{wu2023graph_ood_gnnsafe}, we employ a margin-based score regularization that enforces a separation between ID and exposed OOD nodes, formulated as:
\begin{equation}
\mathcal{L}_{\text{reg}}
=
\frac{\sum_{v_i \in \mathcal{V}_{id}}
[\sigma(s_i - \Delta_1)]^2}{|\mathcal{V}_{id}|} +
\frac{\sum_{v_j \in \mathcal{V}_{\text{exp}}}
[\sigma(\Delta_2 - s_j)]^2}{|\mathcal{V}_{\text{exp}}|},
\end{equation}
where $\sigma$ denotes the ReLU function, $\Delta_1$ and $\Delta_2$ are predefined score margins for ID and OOD nodes, respectively.
This regularization encourages low OOD scores for ID nodes while pushing exposed OOD nodes toward higher scores.

Finally, the optimization objective can be calculated as follows:
\begin{equation}
\label{eq: integration}
\mathcal{L}
=
\mathcal{L}_{\text{sup}}
+\beta\, \mathcal{L}_{\text{reg}},
\end{equation}
where $\mathcal{L}_{\text{sup}}$ denotes the downstream task loss, $\mathcal{L}_{\text{reg}}$ is the proposed OOD exposure regularization, and $\beta$ controls the trade-off. As $\mathcal{L}_{\text{reg}}$ is applied at the score level without imposing architectural constraints on $f_\theta$, \textsc{LG-Plug} can be seamlessly integrated into topology-based graph OOD detectors while preserving their original pipelines. The overall procedure is summarized in Algorithm~\ref{alg: lg_plug}.

   \begin{algorithm}[htbp]\fontsize{8pt}{4pt}\selectfont
    \DontPrintSemicolon
    \SetAlgoLined
    \caption{Overall Procedure of \textsc{LG-Plug}}
    \label{alg: lg_plug}
    \SetKwFunction{FServer}{S\_Exec}
    \SetKwFunction{FClient}{C\_Exec}
    \SetKwProg{Fn}{Function}{:}{end}

    \KwInput{Text-attributed graph $G$, ID categories $\mathcal{Y}_{id}$, parameter weights of graph encoder $\theta_{GE}$ and text encoder $\theta_{TE}$, number of clusters $M$, LLM query batch size $b$, LLM trial times $T$, codebook filtering top-$K$.}
    \tcc{Topology-text representation alignment}
    {$\theta_{GE}, \theta_{TE} \leftarrow $ parameter initialization;} \\
    \While{not converged}{
        {sample batches of nodes and their textual descriptions in $G$;}\\
        \For{each batch}{
            compute graph embedding $Z$ via Eq.~(\ref{eq: graph embedding});\\
            compute text embedding $H$ via Eq.~(\ref{eq: text embedding});\\
            compute node-level alignment loss $\mathcal{L}_\text{node}$ via Eq.~(\ref{eq: node loss});\\
            compute edge-level alignment loss $\mathcal{L}_\text{edge}$ via Eq.~(\ref{eq: edge loss});\\
            update $\theta_{GE}, \theta_{TE}$ with backpropagation via Eq.~(\ref{eq: ttra obj});
        }
    }

    \tcc{Consensus-driven OOD Exposure with LLM}
    {obtain node clusters $\{\mathcal{C}_m\}_{m=1}^M$ via Eq.~(\ref{eq: clustering})};\\
    {$\mathcal{V}_{\text{exp}} \leftarrow \emptyset$ (OOD exposure initialization);}\\
    \For{each cluster $\mathcal{C}_m$ in $\{\mathcal{C}_m\}_{m=1}^M$}{
        compute near-centroid cluster $\mathcal{\hat{C}}_m$ via Eq.~(\ref{eq: near-centroid cluster});\\
        category codebook initialization with $\mathcal{Y}_{id}$ via Eq.~(\ref{eq: codebook init});\\
        
        \For{each LLM trail $t$ in $1,...,T$}{
            sample batches of nodes $\mathcal{R}_m$ in $\mathcal{\hat{C}}_m$;\\
            \For{each node $v_i$ in $\mathcal{R}_m$}{
                assign a category $\hat{c}_i$ for node $v_i$ via Eq.~(\ref{eq: llm query});\\
                update the category codebook $\mathcal{B}_m$ via Eq.~(\ref{eq: update codebook});\\
            }
            \If{$|\{\hat{c}_i\}_{i=1}^b| = 1$}{
                \textbf{break}; \tcp*{the batch shares the same category}
            }
        }
        obtain top-$K$ categories $\mathcal{Y}_{top}$ via Eq.~(\ref{eq: top-K items});\\
        \If{$\mathcal{Y}_{top} \cap\mathcal{Y}_{id} = \emptyset$}{
            update OOD exposure set $\mathcal{V}_{exp}$ via Eq.~(\ref{eq: ood exposure});
        }

    }
    \tcc{Score Regularization via OOD Exposure}
    integrate \textsc{LG-Plug} with topology-driven OOD methods via Eq.~(\ref{eq: integration}).
    \end{algorithm}

\begin{table*}[t]
\centering
\caption{\textbf{OOD detection performance on six TAG datasets.} We report FPR95 and AUROC, and all metrics are percentages. $\uparrow/\downarrow$ indicate higher/lower is better. The best, second best and third best results are highlighted in \textcolor{darkred}{\textbf{red}}, \textcolor{royalblue}{\textbf{blue}} and \textcolor{orange}{\textbf{orange}}, respectively.}
\label{tab: main_results}
\footnotesize 
\renewcommand{\arraystretch}{1.1}

\resizebox{\linewidth}{35mm}{  %
\setlength{\tabcolsep}{1.2mm}{
\begin{tabular}{l!{\vrule width 0.1pt}cc!{\vrule width 0.1pt}cc!{\vrule width 0.1pt}cc!{\vrule width 0.1pt}cc!{\vrule width 0.1pt}cc!{\vrule width 0.1pt}cc}
\hline\thickhline
\rowcolor{gray!20}
 & \multicolumn{2}{c!{\vrule width 0.1pt}}{\textbf{Cora}} & \multicolumn{2}{c!{\vrule width 0.1pt}}{\textbf{CiteSeer}} & \multicolumn{2}{c!{\vrule width 0.1pt}}{\textbf{PubMed}} & \multicolumn{2}{c!{\vrule width 0.1pt}}{\textbf{WikiCS}} & \multicolumn{2}{c!{\vrule width 0.1pt}}{\textbf{Books-History}} & \multicolumn{2}{c}{\textbf{ogbn-arxiv}} \\
\cline{2-13}
\rowcolor{gray!20}
\multirow{-2}{*}{\textbf{Methods}} & FPR95$\downarrow$ & AUROC$\uparrow$ & FPR95$\downarrow$ & AUROC$\uparrow$ & FPR95$\downarrow$ & AUROC$\uparrow$ & FPR95$\downarrow$ & AUROC$\uparrow$ & FPR95$\downarrow$ & AUROC$\uparrow$ & FPR95$\downarrow$ & AUROC$\uparrow$ \\
\hline
\rowcolor{gray!80}
\multicolumn{13}{c}{\textit{\textbf{\textcolor{white}{Classical OOD Methods}}}} \\
MSP~\cite{Hendrycks2017MSP} & 60.99 & 80.92 & 73.80 & 76.03 & 90.82 & 44.07 & 65.66 & 79.51 & 57.06 & 80.81 & 68.92 & 81.05 \\
\rowcolor{gray!10} ODIN~\cite{liang2018odin} & 56.83 & 82.87 & 70.26 & 80.57 & 90.82 & 44.07 & 52.73 & 85.93 & 54.21 & 84.03 & 58.69 & 87.62 \\
Mahalanobis~\cite{lee2018mahalanobis} & 73.51 & 70.91 & 64.37 & 81.12 & 80.50 & 69.70 & 78.48 & 69.24 & 89.97 & 67.46 & 69.22 & 81.44 \\

\rowcolor{gray!80}
\multicolumn{13}{c}{\textit{\textbf{\textcolor{white}{Topology-driven Graph OOD Methods}}}} \\
\textsc{GPN}~\cite{stadler2021graph_ood_gpn} & 63.64 & 85.65 & 71.55 & 76.89 & 63.77 & 86.75 & 55.17 & 84.65 & 51.02 & 80.74 & 63.70 & 83.46 \\

\rowcolor{gray!10} \textsc{GNNSafe}~\cite{wu2023graph_ood_gnnsafe} & 50.47 & 89.80 & 69.77 & 79.01 & 67.28 & 88.42 & 53.51 & 86.20 & 47.72 & 82.81 & 60.84 & 87.98 \\

\textsc{NodeSafe}~\cite{yang2025graph_ood_nodesafe} & 44.21 & 90.40 & 69.22 & 80.32 & 50.35 & \textcolor{orange}{\textbf{92.16}} & 58.96 & 85.54 & 50.81 & 84.21 & 44.58 & 88.64 \\

\rowcolor{gray!10} \textsc{GOLD}~\cite{wang2025graph_ood_gold} & \textcolor{orange}{\textbf{30.35}} & 92.48 & 67.14 & \textcolor{orange}{\textbf{86.06}} & 48.21 & 91.98 & 34.41 & 85.28 & 45.90 & 81.61 & 38.82 & 90.49 \\

\textsc{GRASP}~\cite{ma2024graph_ood_grasp} & 32.82 & \textcolor{orange}{\textbf{93.96}} & 64.62 & 83.53 & \textcolor{orange}{\textbf{35.86}} & 88.08 & \textcolor{orange}{\textbf{20.32}} & \textcolor{orange}{\textbf{93.53}} & 43.03 & 85.38 & \textcolor{orange}{\textbf{27.36}} & \textcolor{orange}{\textbf{92.61}} \\
\hline

\rowcolor{gray!80}
\multicolumn{13}{c}{\textit{\textbf{\textcolor{white}{LLM-based Graph OOD Methods}}}} \\
LLMGuard~\cite{lv2025tag_ood_llm_llmguard} & 42.36 & 89.66 & \textcolor{orange}{\textbf{63.80}} & 81.52  & 58.44 & 84.27 & 46.42 & 86.20 & \textcolor{orange}{\textbf{41.25}} & 82.61 & 52.45 & 82.44 \\
\rowcolor{gray!10} GLIP-OOD~\cite{xu2025zero_shot_tag_ood_glip} & 48.32 & 90.60 & 66.43 & 78.55 & 65.39 & 86.62 & 55.38 & 84.10 & 52.40 & \textcolor{royalblue}{\textbf{85.15}} & 57.17 & 84.39 \\
GOE-LLM~\cite{xu2025goe_llm} & 47.28 & 89.44 & 67.52 & 79.26 & 66.38 & 86.10 & 58.30 & 78.42 &   54.12 & 81.37 & 59.20 & 90.33 \\
\hline
\rowcolor{gray!80}
\multicolumn{13}{c}{\textit{\textbf{\textcolor{white}{Topology-driven Graph OOD Methods Integrated with \textsc{LG-Plug} }}}} \\
 \textsc{LG-Plug} + \textsc{GNNSafe} & \textcolor{royalblue}{\textbf{23.60}} & \textcolor{royalblue}{\textbf{95.24}}   &  \textcolor{royalblue}{\textbf{56.25}} & \textcolor{royalblue}{\textbf{88.82}} & \textcolor{royalblue}{\textbf{27.56}} & \textcolor{royalblue}{\textbf{92.74}} & \textcolor{royalblue}{\textbf{15.56}} & \textcolor{royalblue}{\textbf{94.31}} & \textcolor{royalblue}{\textbf{40.18}} & \textcolor{orange}{\textbf{84.33}} & \textcolor{royalblue}{\textbf{21.35}} & \textcolor{royalblue}{\textbf{93.68}} \\
\rowcolor{gray!10} \textsc{LG-Plug} + GRASP &  \textcolor{darkred}{\textbf{17.11}} & \textcolor{darkred}{\textbf{95.87}}  & \textcolor{darkred}{\textbf{48.51}} & \textcolor{darkred}{\textbf{90.42}} & \textcolor{darkred}{\textbf{20.45}} & \textcolor{darkred}{\textbf{94.63}} &  \textcolor{darkred}{\textbf{8.32}} & \textcolor{darkred}{\textbf{97.33}} & \textcolor{darkred}{\textbf{36.03}} & \textcolor{darkred}{\textbf{89.58}} & \textcolor{darkred}{\textbf{16.36}} & \textcolor{darkred}{\textbf{95.61}} \\
\hline\thickhline
\end{tabular}
}}
\end{table*}

\begin{figure*}[t]
    \centering
    \begin{subfigure}{0.19\linewidth}
        \centering
        \includegraphics[width=\linewidth]{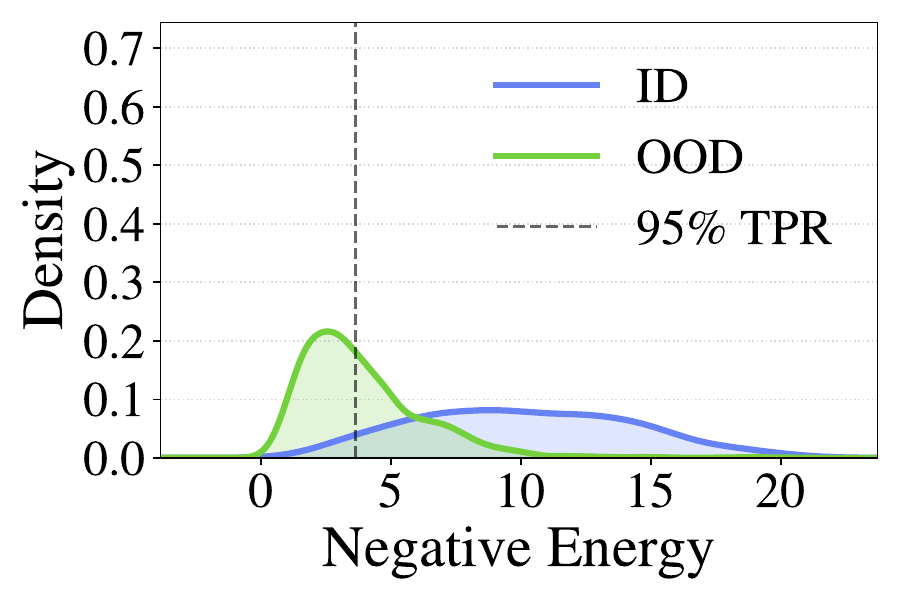}
        \caption{\textsc{GNNSafe}}
    \end{subfigure}\hfill
    \begin{subfigure}{0.19\linewidth}
        \centering
        \includegraphics[width=\linewidth]{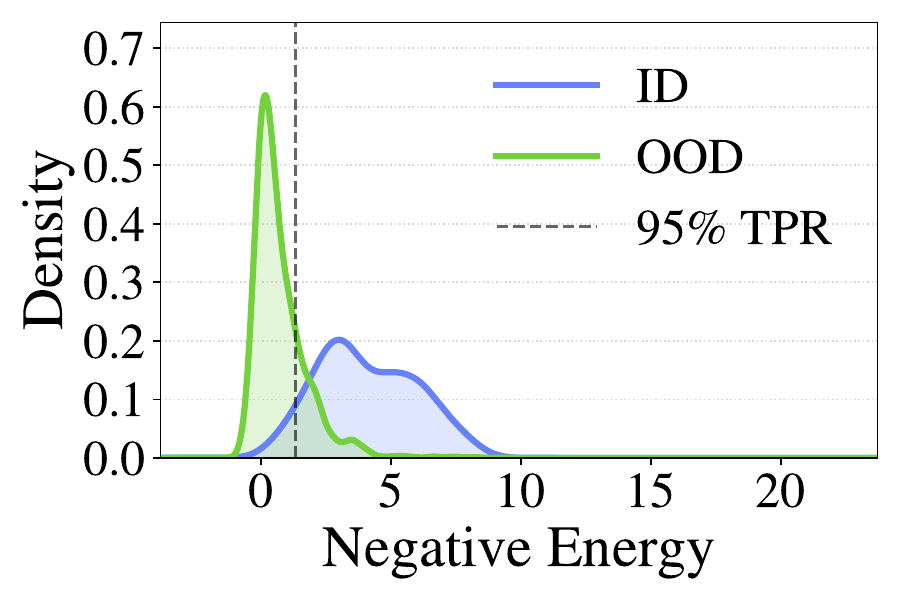}
        \caption{\textsc{LG-Plug} + \textsc{GNNSafe}}
    \end{subfigure}\hfill
    \begin{subfigure}{0.19\linewidth}
        \centering
        \includegraphics[width=\linewidth]{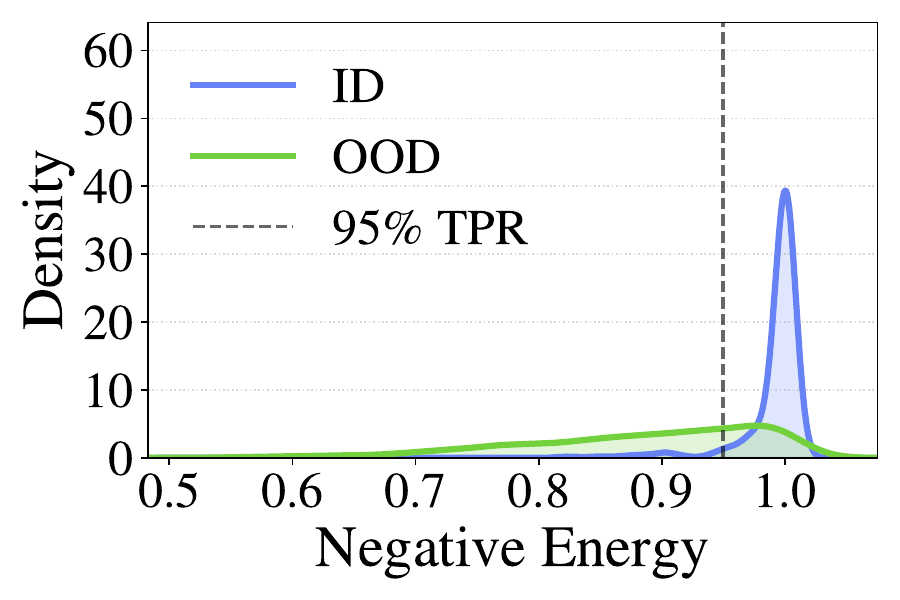}
        \caption{\textsc{GRASP}}
    \end{subfigure}\hfill
    \begin{subfigure}{0.19\linewidth}
        \centering
        \includegraphics[width=\linewidth]{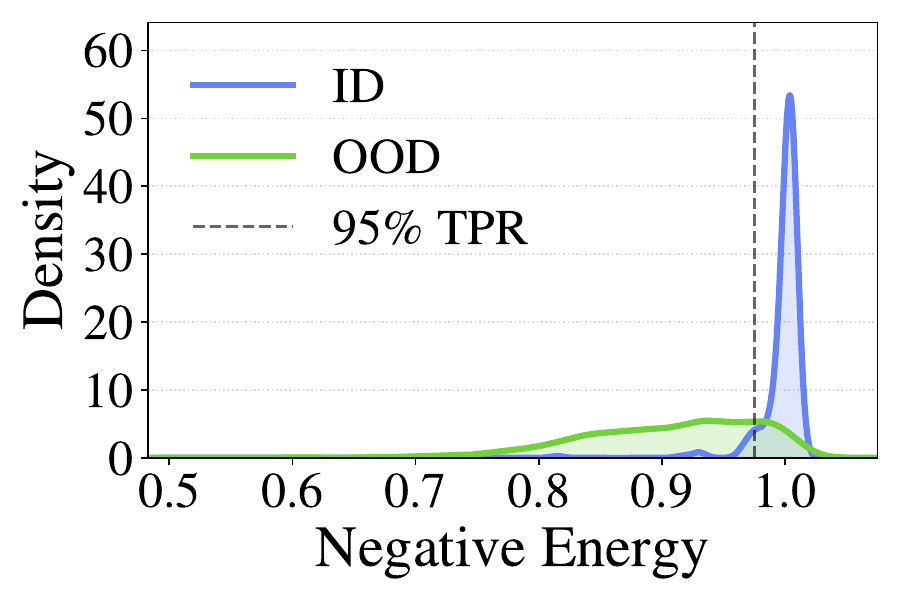}
        \caption{\textsc{LG-Plug} + \textsc{GRASP}}
    \end{subfigure}\hfill
    \begin{subfigure}{0.19\linewidth}
        \centering
        \includegraphics[width=\linewidth]{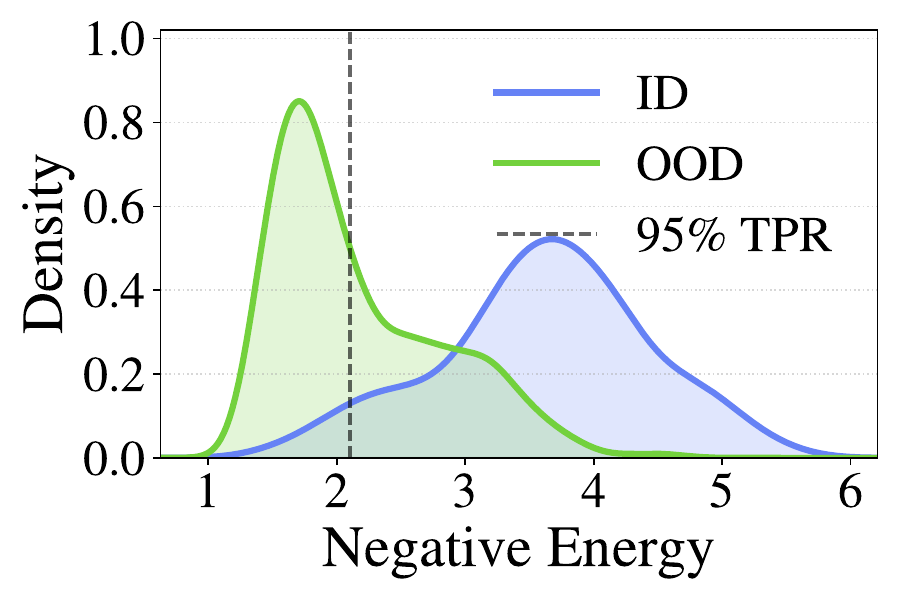}
        \caption{LLMGuard}
    \end{subfigure}

    \caption{Distribution of negative energy scores on the Cora dataset. The dashed line indicates the threshold for a 95\% true positive rate (TPR).}
    \label{fig: density}
\end{figure*}

\section{Experiment}
\label{sec: experiment}
In this section, we present a comprehensive evaluation of \textsc{LG-Plug}.
We introduce the experimental settings (Sec.~\ref{sec: experimental setup}),
and then investigate the following research questions: (1) \textbf{RQ1 \& RQ2}: How effectively does \textsc{LG-Plug} enhance classical and topology-driven detectors, and does it outperform state-of-the-art LLM-based baselines? (Sec.~\ref{sec: performance comparison}). \textbf{RQ3}: What are the individual contributions of each module in \textsc{LG-Plug} to the overall performance (Sec.~\ref{sec: ablation study})? \textbf{RQ4}: How robust is \textsc{LG-Plug} to variations in hyperparameters (Sec.~\ref{sec: sensitivity analysis})? \textbf{RQ5}: Compared to LLM-based graph OOD detection methods, does \textsc{LG-Plug} show superior efficiency (Sec.~\ref{sec: efficiency})? \textbf{RQ6}: How does \textsc{LG-Plug} capture OOD semantics through its exposure (Sec.~\ref{sec: ood exposure quality})?

\subsection{Experimental Setup}
\label{sec: experimental setup}

 \noindent \textbf{Datasets.} We conduct extensive experiments on six public TAG benchmarks varied from different domains (Academic, Wikipedia and E-commerce) and scales (small, medium and large), including Cora, Citeseer, PubMed~\cite{sen2008tag_dataset_cora_citeseer}, WikiCS~\cite{mernyei2020tag_dataset_wiki}, Books-History~\cite{yan2023tag_dataset_history} and ogbn-arxiv~\cite{hu2020ogb}. More details are provided in \cite{LGPLUG2026}.

\vspace{+0.1cm}
\noindent \textbf{Metrics.} We evaluate OOD detection performance using AUROC and FPR95. AUROC measures the area under the ROC curve, where higher values indicate better discrimination, while FPR95 denotes the false positive rate of OOD samples when the ID true positive rate is fixed at 95\%, with lower values being preferable. 

\vspace{+0.1cm}
\noindent \textbf{Baselines.} In this paper, we consider three categories of baselines: (1) \textbf{Classical OOD Methods}, which are originally designed for Euclidean data and adapted to graphs, including MSP~\cite{Hendrycks2017MSP}, ODIN~\cite{liang2018odin}, and Mahalanobis~\cite{lee2018mahalanobis}; (2) \textbf{Topology-driven graph OOD Methods}, which focus on characterize the OODness of a target node through the OODness of its topological neighbors, including GPN~\cite{stadler2021graph_ood_gpn}, \textsc{GNNSafe}~\cite{wu2023graph_ood_gnnsafe}, GRASP~\cite{ma2024graph_ood_grasp}, \textsc{NodeSafe}~\cite{yang2025graph_ood_nodesafe}, and GOLD~\cite{wang2025graph_ood_gold}; (3) \textbf{LLM-based graph OOD Methods}, which are specifically designed for OOD detection on TAGs by utilizing LLM for advanced textual understanding. These methods serve as our main comparison targets, including LLMGuard~\cite{lv2025tag_ood_llm_llmguard}, GLIP-OOD~\cite{xu2025graph_zero_shot_ood_glip}, and GOE-LLM~\cite{xu2025goe_llm}. Further details are presented in \cite{LGPLUG2026}.

\vspace{+0.1cm}
\noindent \textbf{Plug Integration.} As a plug-and-play strategy, the effectiveness of \textsc{LG-Plug} is primarily evaluated by the performance gains obtained when integrating it with existing topology-driven graph OOD detection methods.
To ensure clarity while maintaining generality, we consider two representative integrations: \textbf{\textsc{LG-Plug} + \textsc{GNNSafe}}, and \textbf{\textsc{LG-Plug} + GRASP}.

\subsection{Performance Comparison (RQ1 \& RQ2)}
\label{sec: performance comparison}

To answer \textbf{RQ1} and \textbf{RQ2}, we conduct a comprehensive performance comparison of \textsc{LG-Plug} against multiple categories of baselines across six TAG benchmarks. The results are summarized in Table~\ref{tab: main_results} and Fig.~\ref{fig: density}. We summarize our observations as follows:

\vspace{0.1cm}
\noindent \textbf{Comparison with Classical OOD Methods.} Classical OOD detectors such as MSP, ODIN, and Mahalanobis exhibit the weakest overall performance. This limitation can be mainly attributed to two factors: (1) These methods were originally designed for Euclidean data and primarily rely on node-wise detection performance, thereby failing to capture OOD signals arising from topological dependencies among nodes. (2) They lack adequate modeling of the textual modality in the TAG setting, making them incapable of understanding complex and fine-grained node textual descriptions.

\begin{table*}[t!]
\centering
\captionsetup{font={small,stretch=0.5}}
\caption{Ablation study of the proposed \textsc{LG-Plug} under the \textsc{GNNSafe} framework on four benchmark datasets.}
\label{table: ablation study}
\resizebox{\linewidth}{13mm}{  
\setlength{\tabcolsep}{4.0mm}{
\begin{tabular}{ccc|cc|cc|cc|cc}
\hline\thickhline
\rowcolor{gray!20}
\multicolumn{3}{c|}{\textbf{Ablation Settings}} & \multicolumn{2}{c|}{\textbf{Cora}} & \multicolumn{2}{c|}{\textbf{Citeseer}} & \multicolumn{2}{c}{\textbf{PubMed}} & \multicolumn{2}{c}{\textbf{ogbn-arxiv}}\\
\cline{1-11}
\rowcolor{gray!20}
$\mathcal{L}_\text{node}$ & $\mathcal{L}_\text{edge}$ & \textit{ICF}     & FPR $\downarrow$ & AUROC $\uparrow$ & FPR $\downarrow$ & AUROC $\uparrow$
 & FPR $\downarrow$ & AUROC $\uparrow$ & FPR $\downarrow$ & AUROC $\uparrow$
 \\
\hline
 \multicolumn{3}{c|}{\textsc{GNNSafe} \textbf{(lower bound)}} & 50.47 & 89.80 & 69.77 & 79.01 & 67.28 & 88.42 & 60.84 & 87.98  \\
\rowcolor{gray!20}
   & \checkmark & \checkmark &  \textcolor{orange}{\textbf{33.52}} & \textcolor{orange}{\textbf{91.28}}  & \textcolor{orange}{\textbf{61.35}} & \textcolor{orange}{\textbf{84.20}}  & \textcolor{orange}{\textbf{39.47}}  & \textcolor{royalblue}{\textbf{91.25}} & \textcolor{orange}{\textbf{30.20}} & \textcolor{orange}{\textbf{89.15}}\\
 \checkmark  &  & \checkmark & \textcolor{royalblue}{\textbf{25.46}}   &  \textcolor{royalblue}{\textbf{93.60}} & \textcolor{royalblue}{\textbf{58.33}} & \textcolor{royalblue}{\textbf{85.49}} & \textcolor{royalblue}{\textbf{34.60}}  & \textcolor{orange}{\textbf{90.75}} & \textcolor{royalblue}{\textbf{26.31}} & \textcolor{royalblue}{\textbf{89.74}}\\

\rowcolor{gray!20}
\checkmark & \checkmark &   & 39.45 & 90.42 &  64.48 & 81.20 & 42.14 & 89.75 & 44.20 & 87.22\\
\multicolumn{3}{c|}{\textsc{LG-Plug} + \textsc{GNNSafe} \textbf{(upper bound)}} & \textcolor{darkred}{\textbf{23.60}} & \textcolor{darkred}{\textbf{95.24}} & \textcolor{darkred}{\textbf{56.25}} & \textcolor{darkred}{\textbf{88.82}} & \textcolor{darkred}{\textbf{27.56}}  & \textcolor{darkred}{\textbf{92.74}} &  \textcolor{darkred}{\textbf{21.35}} & \textcolor{darkred}{\textbf{93.68}} \\
\hline\thickhline
\end{tabular}%
}}
\end{table*}

\vspace{0.1cm}
\noindent \textbf{Comparison with Topology-driven Graph OOD Methods.}
The five evaluated topology-driven graph OOD methods consistently outperform classical OOD baselines, demonstrating their effectiveness in leveraging topology-level insights to mitigate node-level detection biases. However, as expected, these methods generally underperform their corresponding LG-Plug-integrated variants (e.g., \textsc{GNNSafe} vs. \textsc{LG-Plug} + \textsc{GNNSafe}, and GRASP vs. \textsc{LG-Plug} + GRASP), due to their limited ability to effectively model the textual modality of nodes in TAG.

\vspace{0.1cm}
\noindent \textbf{Comparison to LLM-based Graph OOD Methods.}
Our observations can be summarized into two key points. 
(1) Overall, LLM-based graph OOD methods achieve performance comparable to topology-driven graph OOD methods. Notably, GRASP, which represents the strongest topology-driven baseline in our evaluation, consistently outperforms all LLM-based graph OOD methods. This suggests that, for OOD detection on TAG, the topological modality constitutes a crucial and non-negligible source of information. In contrast, more advanced modeling of node textual semantics can be viewed as a complementary factor that further improves the upper bound of detection performance. (2) Methods integrated with \textsc{LG-Plug} consistently outperform LLM-based graph OOD methods, indicating the necessity of jointly modeling textual and topological modalities to better capture OODness for TAGs.

\vspace{0.1cm}
\noindent \textbf{OOD Score Visualization.} We visualize the frequency density of OOD scores (i.e., negative energy) produced by five methods (\textsc{GNNSafe}, \textsc{LG-Plug}+\textsc{GNNSafe}, \textsc{GRASP}, \textsc{LG-Plug}+\textsc{GRASP}, and LLMGuard) on the Cora dataset in Fig.~\ref{fig: density}. As a plug-in module, \textsc{LG-Plug} preserves the overall shape of the score distributions induced by topology-driven graph OOD detectors, while substantially reducing the overlap between ID and OOD nodes by introducing reliable and informative OOD exposure. In contrast, although LLMGuard generates pseudo OOD samples using LLMs, these samples exhibit a semantic gap from real graph OOD nodes, resulting in limited separation in score distributions and marginal detection gains.

\subsection{Ablation Study (RQ3)}
\label{sec: ablation study}

To address \textbf{RQ3}, we perform an ablation study to assess the contribution of individual components in \textsc{LG-Plug}. The strategy comprises two key modules: Topology-Text Representation Alignment (\textit{\underline{TTRA}}), which jointly trains graph and text encoders to learn fine-grained node representations and induce initial ID-OOD separation, and Consensus-driven OOD Exposure with LLM (\textit{\underline{COEL}}), which exploits LLMs to select reliable OOD exposures from unlabeled nodes. To evaluate \textit{TTRA}, we ablate its node-level and edge-level alignment objectives ($\mathcal{L}\text{node}$ and $\mathcal{L}\text{edge}$), while for \textit{COEL}, we remove the intra-cluster filtering (\textit{ICF}) mechanism that improves the reliability of selected OOD samples.

All ablation experiments are conducted using \textsc{GNNSafe} as the detector backbone, where the vanilla \textsc{GNNSafe} serves as the lower-bound baseline and the full \textsc{LG-Plug} integrated with \textsc{GNNSafe} acts as the upper-bound reference. Results in Table~\ref{table: ablation study} show that removing any component causes a non-negligible performance drop, indicating that both representation alignment and intra-cluster filtering (\textit{ICF}) are essential to \textsc{LG-Plug}. Moreover, ablating \textit{ICF} consistently leads to the largest degradation, though still outperforming the naive \textsc{GNNSafe}, suggesting that \textit{ICF} is crucial for acquiring high-quality OOD exposure, while the effectiveness of the consensus-driven OOD exposure with LLMs does not solely rely on this mechanism.

\begin{figure}[h]
    \includegraphics[width=0.48\textwidth]{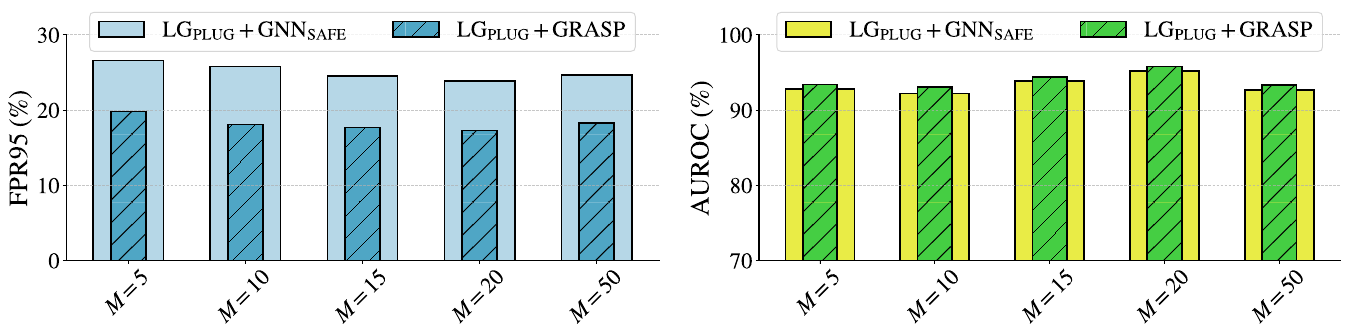}
    \captionsetup{font={small,stretch=1}}
    \caption{Sensitivity analysis for the number of clusters (corresponding to Eq.~(\ref{eq: clustering})) on the Cora dataset.}
    \label{fig: sen_M}
\end{figure}

\subsection{Sensitivity Analysis (RQ4)}
\label{sec: sensitivity analysis}

To answer \textbf{RQ4}, we conduct a sensitivity analysis to examine the robustness of \textsc{LG-Plug} with respect to key hyperparameters, including the number of clusters $M$, the intra-cluster filtering threshold $\rho$, and the top-frequent matching category $K$. The remaining hyperparameter settings are reported in \cite{LGPLUG2026}.

As shown in Fig.~\ref{fig: sen_M}, when the number of clusters varies from 5 to 50, \textsc{LG-Plug} consistently maintains stable performance and continues to improve topology-driven graph OOD detection methods. This observation indicates that \textsc{LG-Plug} is insensitive to variations in the clustering granularity and demonstrates strong robustness across different settings. Consequently, \textsc{LG-Plug} can be reliably deployed without requiring extensive hyperparameter tuning.

\begin{table*}[t]
    \caption{Comprehensive efficiency analysis on the Cora dataset. We evaluate theoretical complexity, empirical time, model parameters (Theoretical and Practical). For simplify, we focus on the language-related computational procedure and parameter counts.}
    \vspace{0.05cm}
    \centering
    \label{tab: complexity}
\resizebox{\linewidth}{10mm}{  
\setlength{\tabcolsep}{8.0mm}{
    \begin{tabular}{c|cc|cc}
    \hline\thickhline
    \rowcolor{gray!20}
         \textbf{Method} & \textbf{Theo. Time} & \textbf{Prac. Time} & \textbf{Theo. Params} & \textbf{Prac. Params} \\
    \hline
         LLMGuard & $\mathcal{O}(C \cdot (K + KL)\cdot T+\Delta_1)$ & 5400s & $\Theta_{Local-LLM}$ & 7B \\
         \rowcolor{gray!20} GLIP-OOD & $\mathcal{O}(N_s \cdot T)$ & 295s & $\Theta_{GFM}$ & 80M \\
         GOE-LLM & $\mathcal{O}(N_s \cdot T)$ & 324s & $\Theta_{TE}$ & 110M \\
         \rowcolor{gray!20} \textsc{LG-Plug} (Ours) & $\mathcal{O}(M \cdot b \cdot T+\Delta_2)$ & \textbf{70s} & $\Theta_{TE}$ & \textbf{63 M}  \\
    \hline\thickhline
    \end{tabular}
    }}
\end{table*}

\begin{figure*}[t]
    \centering
    \begin{subfigure}{0.23\linewidth}
        \centering
        \includegraphics[width=\linewidth]{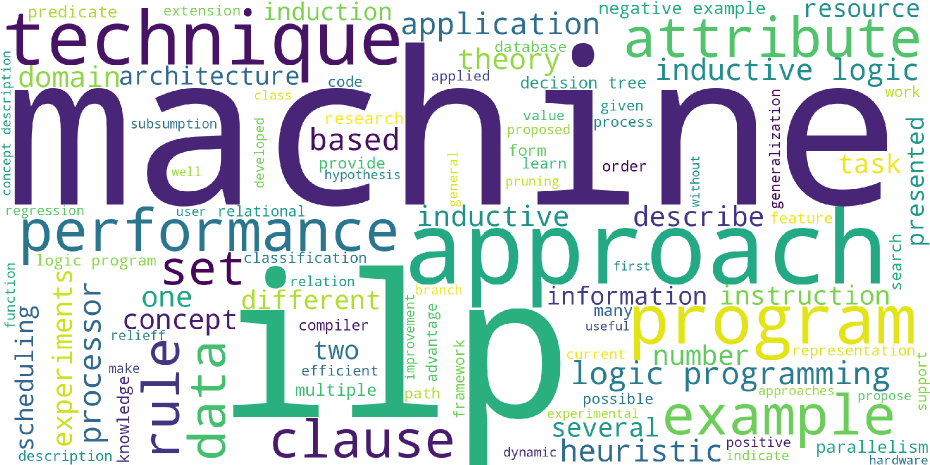}
        \caption{Rule Learning (ID)}
    \end{subfigure}\hfill
    \begin{subfigure}{0.23\linewidth}
        \centering
        \includegraphics[width=\linewidth]{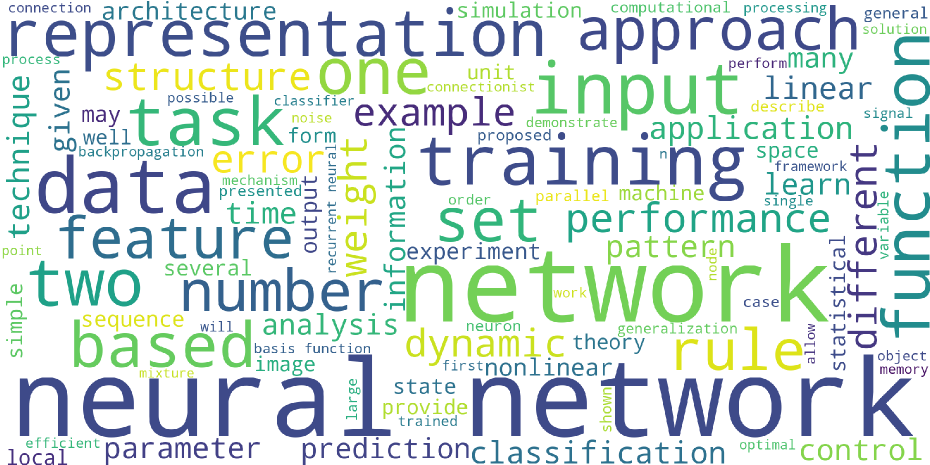}
        \caption{Neural Networks (ID)}
    \end{subfigure}\hfill
    \begin{subfigure}{0.23\linewidth}
        \centering
        \includegraphics[width=\linewidth]{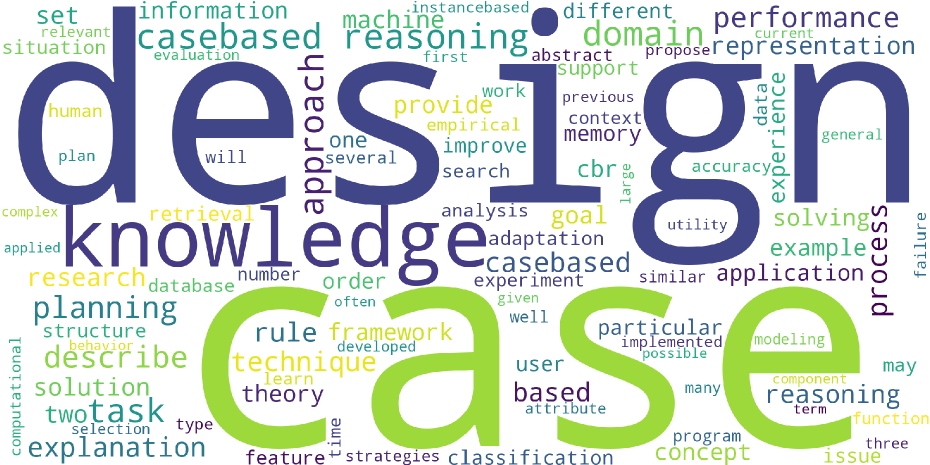}
        \caption{Case Based (ID)}
    \end{subfigure}\hfill
    \begin{subfigure}{0.23\linewidth}
        \centering
        \includegraphics[width=\linewidth]{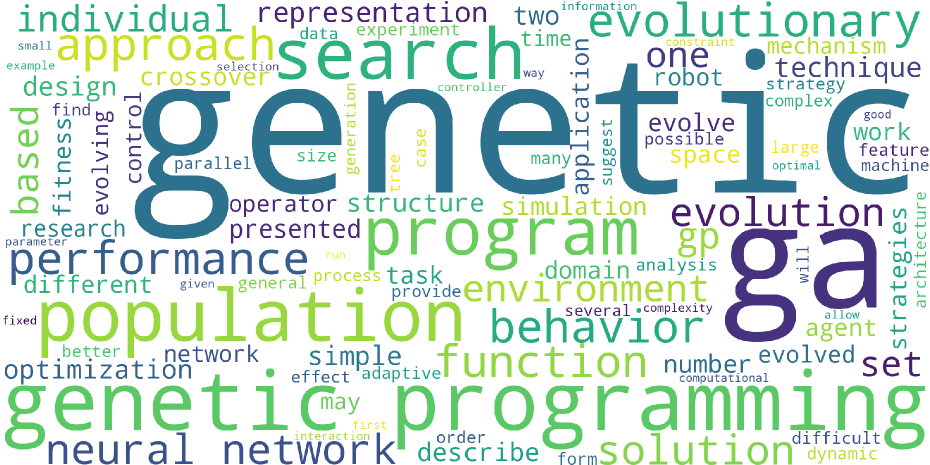}
        \caption{Genetic Algorithms (OOD)}
    \end{subfigure}

    \vspace{0.5em} 

    \begin{subfigure}{0.23\linewidth}
        \centering
        \includegraphics[width=\linewidth]{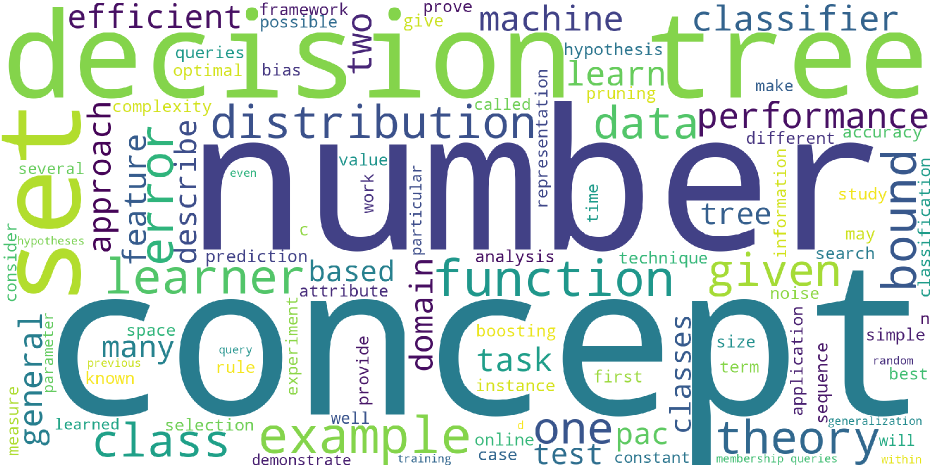}
        \caption{Theory (OOD)}
    \end{subfigure}\hfill
    \begin{subfigure}{0.23\linewidth}
        \centering
        \includegraphics[width=\linewidth]{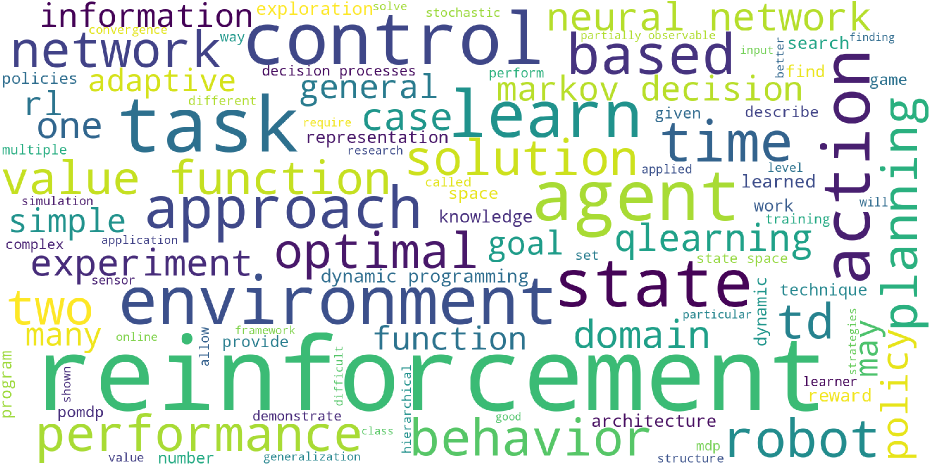}
        \caption{Reinforcement Learning (OOD)}
    \end{subfigure}\hfill
    \begin{subfigure}{0.23\linewidth}
        \centering
        \includegraphics[width=\linewidth]{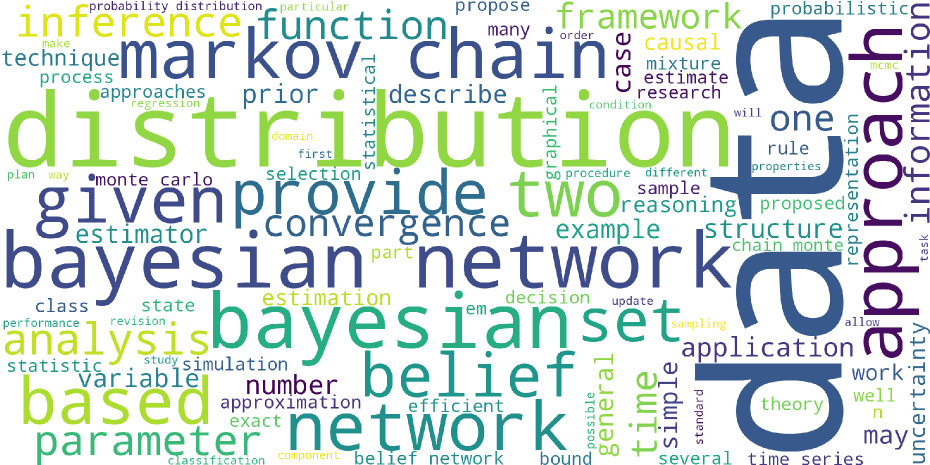}
        \caption{Probabilistic Methods (OOD)}
    \end{subfigure}\hfill
    \begin{subfigure}{0.23\linewidth}
        \centering
        \includegraphics[width=\linewidth]{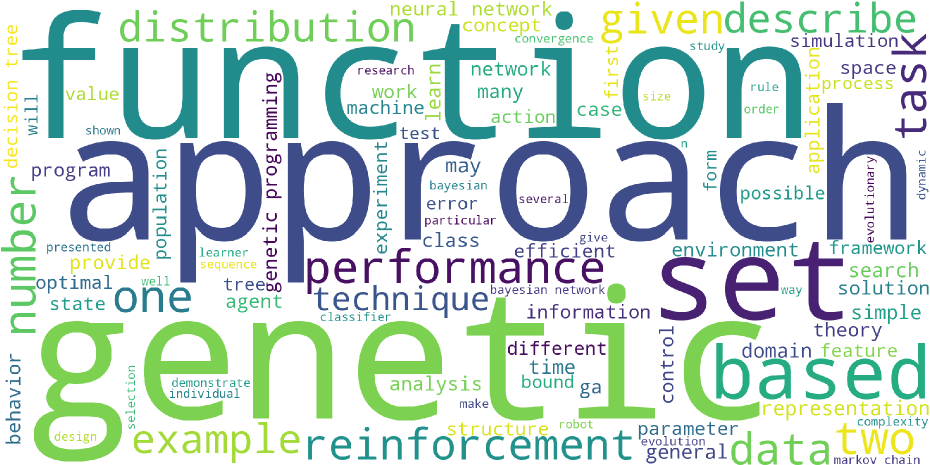}
        \caption{All OOD (Ground-Truth)}
    \end{subfigure}

    \vspace{0.5em} 

    \begin{subfigure}{0.23\linewidth}
        \centering
        \includegraphics[width=\linewidth]{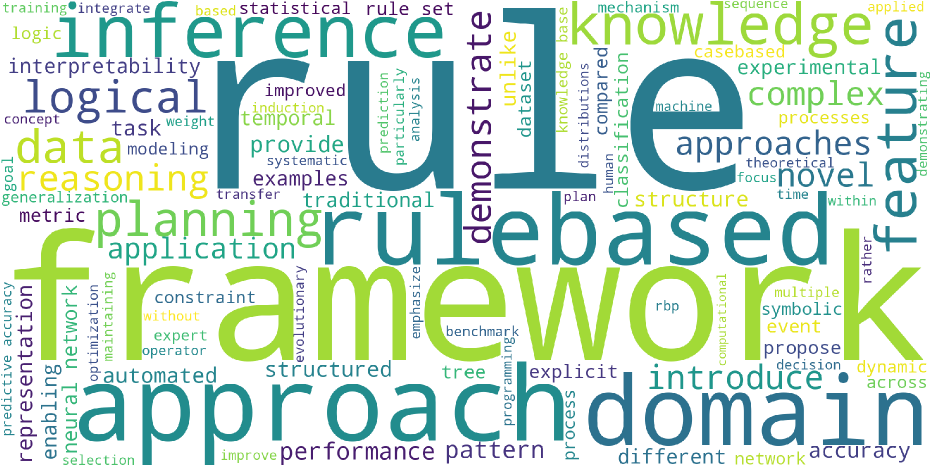}
        \caption{LLMGuard OOD Exposure}
    \end{subfigure}\hfill
    \begin{subfigure}{0.23\linewidth}
        \centering
        \includegraphics[width=\linewidth]{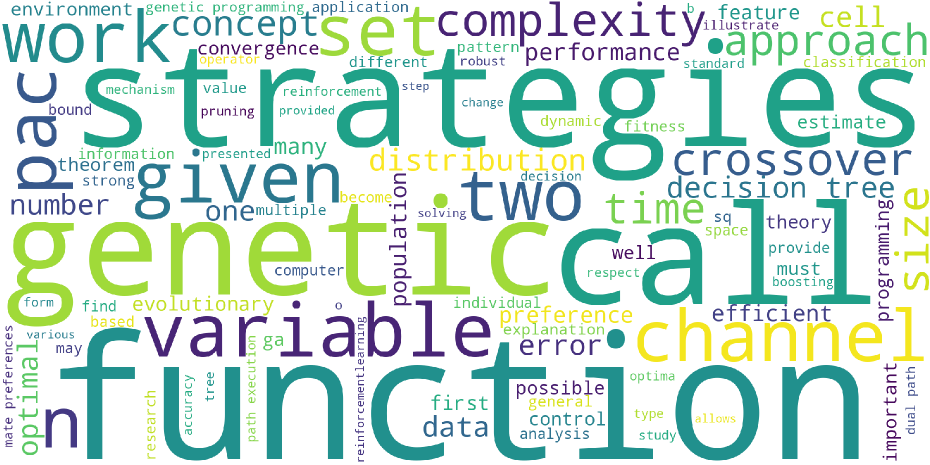}
        \caption{GLIP-OOD OOD Exposure}
    \end{subfigure}\hfill
    \begin{subfigure}{0.23\linewidth}
        \centering
        \includegraphics[width=\linewidth]{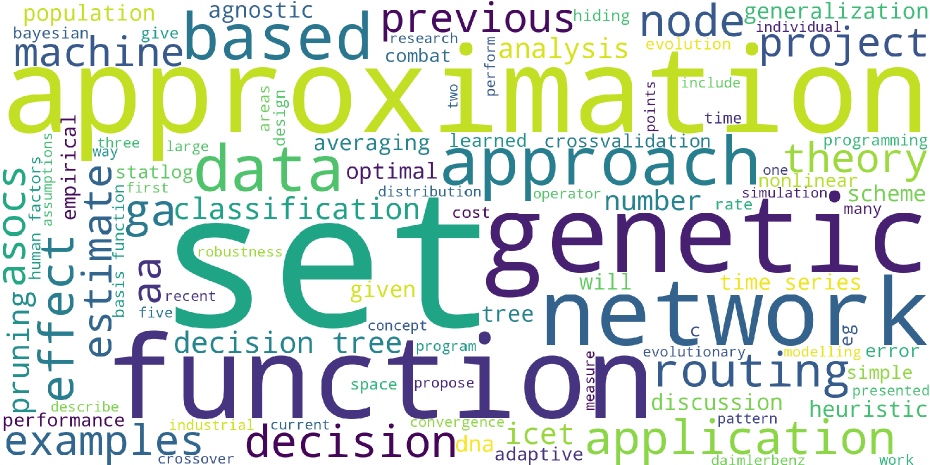}
        \caption{GOE-LLM OOD Exposure}
    \end{subfigure}\hfill
    \begin{subfigure}{0.23\linewidth}
        \centering
        \includegraphics[width=\linewidth]{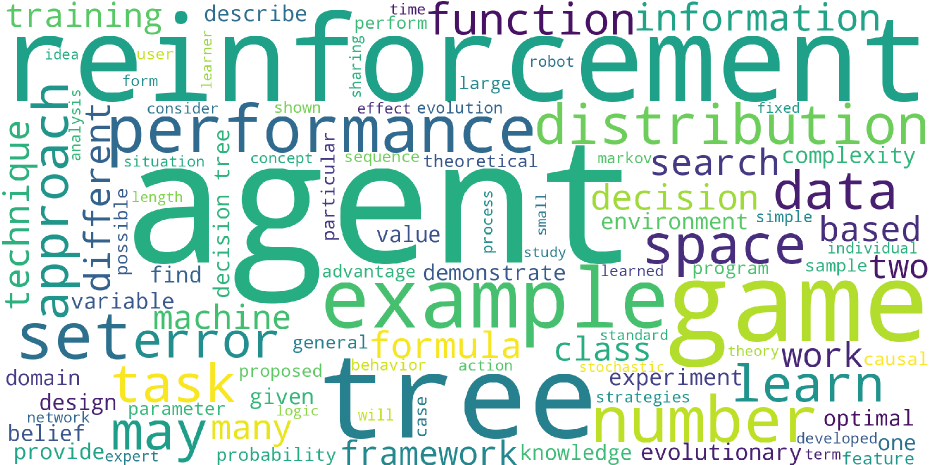}
        \caption{\textsc{LG-Plug} OOD Exposure (Ours)}
    \end{subfigure}

    \caption{Word cloud visualizations on the Cora dataset. (a)-(g) show ground-truth ID and OOD categories, while (h) displays all the OOD nodes. (i)-(l) denotes OOD exposure from our \textsc{LG-Plug} and baselines. Stopwords from both WordCloud’s built-in list and our custom list (\cite{LGPLUG2026}) were removed for clarity.}
    \label{fig: word cloud}
\end{figure*}

\subsection{Efficiency Analysis (RQ5)}
\label{sec: efficiency}

To answer \textbf{RQ5}, we evaluate the efficiency of \textsc{LG-Plug} and representative LLM-based graph OOD detection methods from three aspects: computational complexity, parameter size, and LLM token cost. \textsc{LG-Plug} is integrated with \textsc{GNNSafe}. We first present a theoretical complexity analysis in Table~\ref{tab: complexity}.

\vspace{0.1cm}
\noindent \textbf{Time and Parameters.} Let $N_s$ denote the number of randomly sampled unlabeled nodes, $C$ the number of ID classes, $D$ the feature dimension, and $T$ the cost of a single LLM query. \textbf{(1) LLMGuard.}
LLMGuard incurs a complexity of $\mathcal{O}\!\left(C \cdot (K + K L) \cdot T\right)$, where $K$ is the number of synthetic nodes generated per class and $L$ denotes the number of structure-enhanced generation steps for synthetic OOD nodes. In addition, LLMGuard requires instruction fine-tuning to train a local LLM-based edge predictor, introducing an extra overhead denoted by $\Delta_1$. Its parameter cost is dominated by the local LLM Vicuna-v1.5 with 7B parameters. \textbf{(2) GLIP-OOD and GOE-LLM.}
Both methods rely on random sampling for LLM-based annotation, resulting in a complexity of $\mathcal{O}(N_s \cdot T)$. As the graph scale and the number of semantic categories increase, a large $N_s$ is typically required to ensure sufficient coverage of diverse OOD categories, leading to substantial LLM query costs. The primary parameters of GLIP-OOD come from the text encoder SentenceBERT (80M parameters), while GOE-LLM relies on the graph foundation model GraphCLIP~\cite{zhu2025graphclip} with 110M parameters. \textbf{(3) \textsc{LG-Plug}.} In contrast, \textsc{LG-Plug} adopts a consensus-driven cluster sampling strategy, achieving a complexity of $\mathcal{O}(M \cdot b \cdot T + \Delta_2)$, where $M$ denotes the number of cluster centers and $b$ is the batch size used for consensus verification. \textsc{LG-Plug} additionally requires topology-text representation alignment, incurring an overhead denoted by $\Delta_2$. Owing to the alignment and intra-cluster filtering mechanisms, \textsc{LG-Plug} achieves strong performance with small $M$ and $b$ (typically $M \leq 40$ and $b \leq 5$). Since $M \cdot b \ll N_s$, \textsc{LG-Plug} effectively decouples the LLM query cost from the graph size. Its parameter cost is mainly attributed to the text encoder, which contains 63M parameters.

\subsection{OOD Exposure Quality Investigation (RQ6)}
\label{sec: ood exposure quality}

To address \textbf{RQ6}, we perform a detailed analysis of the \textsc{LG-Plug} mechanism by visualizing its OOD exposure through word clouds, as shown in Fig.~\ref{fig: word cloud}. The visualizations reveal that \textsc{LG-Plug} captures high-frequency semantics across all four OOD categories, effectively covering their distinctive concepts (e.g., ``\textit{evolutionary}'' for ``Genetic Algorithms'', ``\textit{theoretical}'' for ``Theory'', ``\textit{reinforcement}'' for ``Reinforcement Learning'', and ``\textit{markov}'' for ``Probabilistic Methods''). Moreover, there is minimal overlap with ID category semantics, indicating that the OOD exposure is both reliable and informative. In contrast, word clouds generated by existing LLM-based graph OOD baselines exhibit either a semantic shift away from true OOD concepts or an infusion of ID noise.

\section*{Conclusion}
\label{sec: conclusion}

In this paper, we investigate OOD detection for text-attributed graphs and identify a key gap between topology-driven methods, which underexploit rich semantic information, and recent LLM-based approaches, which often suffer from unreliable or weakly informative synthesized OOD priors, high computational and token costs, and limited compatibility with topology-driven detectors. To bridge this gap, we propose \textsc{LG-Plug}, a lightweight plug-and-play strategy that injects LLM-guided semantic supervision into existing topology-driven OOD detectors without architectural modification. Experiments on six benchmark datasets show that \textsc{LG-Plug} consistently improves multiple detectors while remaining efficient. Future work includes extending \textsc{LG-Plug} to dynamic or open-world graph settings and exploring adaptive prompting strategies to enhance the reliability of LLM-guided OOD exposure.

\section*{Acknowledgments}
This work was supported by the Shenzhen Science and Technology Program (Grant No. KJZD20230923113901004), the National Natural Science Foundation of China (Grant Nos. 62572501 and 62502551), the Pearl River Talent Recruitment Program (Grant No. 2024QN11X150), and the Guangdong Basic and Applied Basic Research Foundation (Grant No. 2026A1515030048).

\bibliographystyle{ACM-Reference-Format}
\balance
\bibliography{OOD}

\end{document}